\documentclass[10pt,journal]{IEEEtran}

\usepackage{graphicx}
\usepackage{amsmath}
\usepackage{amssymb}
\usepackage{booktabs}
\usepackage{algorithm}
\usepackage{algpseudocode}
\usepackage{listings}

\usepackage[table]{xcolor}
\usepackage{xcolor}
\usepackage{gensymb}
\usepackage{todonotes}

\definecolor{codegreen}{rgb}{0,0.6,0}
\definecolor{codegray}{rgb}{0.5,0.5,0.5}
\definecolor{codepurple}{rgb}{0.58,0,0.82}
\definecolor{backcolour}{rgb}{0.95,0.95,0.92}

\lstdefinestyle{mystyle}{
    commentstyle=\color{codegreen},
    keywordstyle=\color{magenta},
    numberstyle=\tiny\color{codegray},
    stringstyle=\color{codepurple},
    basicstyle=\ttfamily\footnotesize,
    breakatwhitespace=false,         
    breaklines=true,                 
    captionpos=b,                    
    keepspaces=true,                 
    numbers=left,                    
    numbersep=5pt,                  
    showspaces=false,                
    showstringspaces=false,
    showtabs=false,                  
    tabsize=2
}

\usepackage[pagebackref,breaklinks,colorlinks]{hyperref}

\begin{document}


\title{Learning Better Contrastive View from Radiologist's Gaze}

\author{Sheng Wang, Zixu Zhuang, Xi Ouyang, Lichi Zhang, Zheren Li, Chong Ma,\\ Tianming Liu, Dinggang Shen, \IEEEmembership{Fellow, IEEE}, Qian Wang
\thanks{Sheng Wang, Zixu Zhuang, Lichi Zhang and Zheren Li are with the School of Biomedical Engineering, Shanghai Jiao Tong University, Shanghai, China. E-mail: \{wsheng, zixu.zhuang, lichizhang, lizheren\}@sjtu.edu.cn.}
\thanks{Tianming Liu is with the Department of Computer Science, University of Georgia, GA, USA. E-mail: tliu@cs.uga.edu.}
\thanks{Chong Ma is with the School of Automation, Northwestern Polytechnical University, Xi'an, China. E-mail: mc-npu@mail.nwpu.edu.cn.}
\thanks{Dinggang Shen and Qian Wang are with the School of Biomedical Engineering, ShanghaiTech University, Shanghai, China. E-mail:\{dgshen, wangqian2\}@shanghaitech.edu.cn. 
Xi Ouyang, Zheren Li and Dinggang Shen are with the Department of Research and Development, Shanghai United Imaging Intelligence Co., Ltd., Shanghai, China. E-mail:xi.ouyang@uii-ai.com}}


\maketitle

\begin{abstract}
Recent self-supervised contrastive learning methods greatly benefit from the Siamese structure that aims to minimizing distances between positive pairs. 
These methods usually apply random data augmentation to input images, expecting the augmented views of the same images to be similar and positively paired. 
However, random augmentation may overlook image semantic information and degrade the quality of augmented views in contrastive learning.
This issue becomes more challenging in medical images since the abnormalities related to diseases can be tiny, and are easy to be corrupted (e.g., being cropped out) in the current scheme of random augmentation.
In this work, we first demonstrate that, for widely-used X-ray images, the conventional augmentation prevalent in contrastive pre-training can affect the performance of the downstream diagnosis or classification tasks.
Then, we propose a novel augmentation method, i.e., FocusContrast, to learn from radiologists' gaze in diagnosis and generate contrastive views for medical images with guidance from radiologists' visual attention. 
Specifically, we track the gaze movement of radiologists and model their visual attention when reading to diagnose X-ray images.
The learned model can predict visual attention of the radiologists given a new input image, and further guide the attention-aware augmentation that hardly neglects the disease-related abnormalities.
As a plug-and-play and framework-agnostic module, FocusContrast consistently improves state-of-the-art contrastive learning methods of SimCLR, MoCo, and BYOL by 4.0$\sim$7.0\% in classification accuracy on a knee X-ray dataset.
\end{abstract}

\section{Introduction}


\label{sec:introduction}
Self-supervised contrastive learning has seen great progress recently. It has begun to match supervised pre-training in performance on several downstream tasks~\cite{chen2020simple,he2020momentum,grill2020bootstrap}. 

The basic idea of contrastive learning is to encourage the representations from the positive pairs of the samples to be close in the embedding space, while the representations from the negative pairs are pushed apart. 
Here, two views that are transformed or augmented from the same training sample are usually defined as a positive pair, while two views from different training samples are defined as a negative pair.

The generation of positive pairs, which is essential for the quality of the learned representations, is a major focus in contrastive learning.~\cite{grill2020bootstrap,chen2021exploring,caron2021emerging}. 
A common approach in computer vision generates the positive pair by randomly augmenting a certain image twice, e.g., random cropping, color distorting, rotating, cutout, adding noise, and so on~\cite{zhong2020random,chen2020improved}.

\begin{figure}[t]
    \centering
    \includegraphics[width=0.46\textwidth]{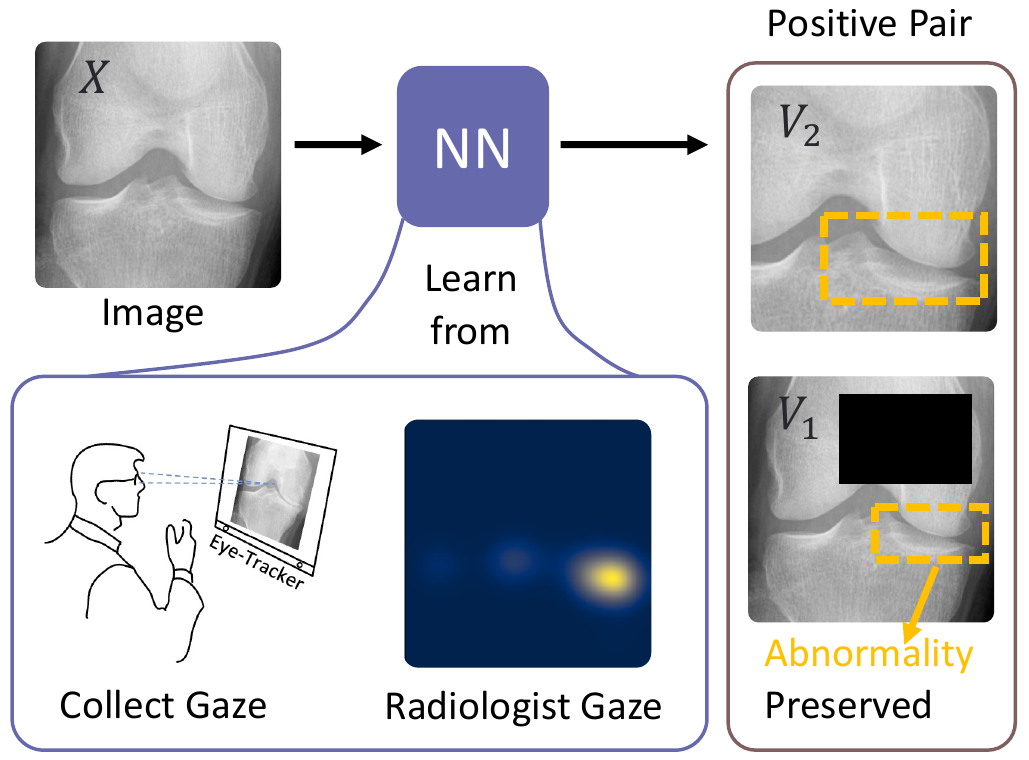}
    \caption{We propose to learn and generate positive contrastive views from the radiologist's gaze. Our proposed framework can preserve highly informative area (e.g., the abnormality in knee OA diagnosis, as highlighted by the yellow boxes) throughout image augmentation, which may be lost by conventional random augmentation.}
    \label{fig:cover_image}
\end{figure}

However, because this process does not take image semantics into account, it sometimes misses the regions of interest that are essential to the subsequent tasks. 
For example, the abnormality related to knee osteoarthritis (OA) diagnosis in Fig.~\ref{fig:cover_image} (highlighted by the yellow box) may be easily lost during the conventional scheme of random cropping or cutout.

\begin{figure*}[t]
    \centering
    \includegraphics[width=0.99\textwidth]{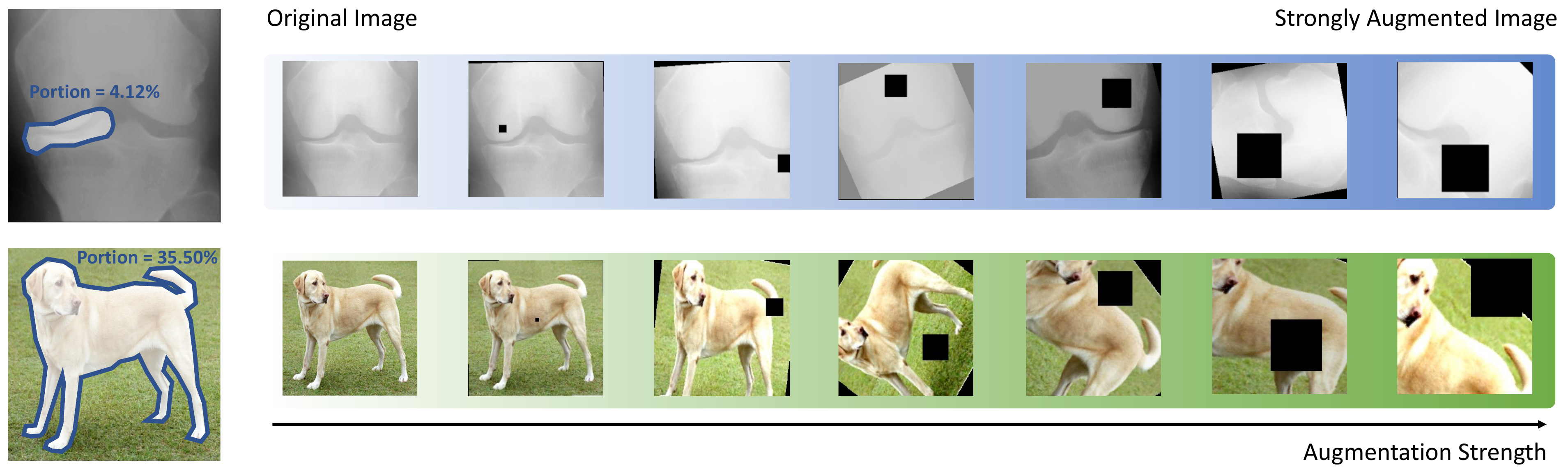}
    \caption{Improper contrastive views in medical image caused by strong random augmentation. Here we take knee X-ray as an example. On the left, one can observe that the decision-related abnormality can be very subtle with a small number of pixels. In comparison with natural images, the object of interest usually occupies a large number of pixels. We also demonstrate increasing aggressiveness of the augmentation applied to the knee X-ray image and the dog image. When the X-ray image is strongly augmented, the semantic cues (i.e., abnormalities) are hard to preserve and recognize, implying that the augmented images may deviate significantly from the subsequent task of diagnosis. In comparison, when the dog image is heavily augmented, the semantic cues are much easier to distinguish and the classification task can be accomplished accordingly.}
    
    \label{fig:hypothesis}
\end{figure*}

There are recent debates about how to design the augmentation strategies and how to control their strength~\cite{tian2020what,xiao2020should,dangovski2021equivariant}. 
For example, 2$\times$ random resized cropping is regarded as a stronger augmentation than 1.2$\times$ random resized cropping, while an excessively strong augmentation may not always yield a good outcome. 
And the augmentation strength should be emphasized especially for some medical image scenarios.
As shown in Fig. \ref{fig:hypothesis}, the visual cues related to the diseased abnormality in the X-ray image may only occupy a tiny number of pixels (e.g., 4.12\% of the image area for the abnormality of femur-tibia space narrowing), which are however expected to provide major information to support clinical decision or diagnosis.
These few pixels are likely to be removed especially when the augmentation strength becomes excessive.
And the resulting positive pairs may corrupt the contrastive representations learned from the images, since the visual cues for subsequent diagnosis are no longer available.
On the contrary, a natural image depicting a dog (with 35.50\% foreground pixels in Fig. \ref{fig:hypothesis}) is less susceptible to the distortion induced by image augmentation. That is, given similar strong image augmentation (at the right-most of the figure), one may still distinguish that the image is for a dog rather than other objects.

Furthermore, this problem may become more challenging due to the limited size of medical image data. 
The positive pairs generated by improper augmentation contribute to incorrect labels (or noisy labels in machine learning literature) during the contrastive pre-training process.
These errors or noisy labels can be addressed with bigger datasets~\cite{rolnick2017deep} and more network parameters~\cite{brown2020language,he2021masked,li2021domain}.
However, it is not always the case for medical images where the size of the dataset is relatively small and the application is complicated. 


Contrastive learning does not show ``close-to-supervised" performance on medical images as it does on natural images, which is partially due to the above-mentioned reasons. 
To evaluate the representation power and the contrastive learning performance, the linear probing protocol is widely used~\cite{chen2020simple,he2020momentum}. That is, a simple linear classifier is trained by freezing the backbone pre-trained through contrastive learning, and the accuracy attained by the linear classifier is used as a measure for the quality of the self-supervised contrastive representations.
With the linear probing protocol, there typically exists a performance gap between the above self-supervised learning paradigm and the conventional fully-supervised learning (i.e., by the end-to-end training of all the parameters of the entire network, instead of only the final linear classifier). The gap is noticeably bigger on medical images, very different from natural images.
For example, the performance difference is about 5\% when applying Momentum Contrast (MoCo) to ImageNet~\cite{he2020momentum};
however, it expands to nearly 10\% when MoCo is applied to CheXpert (i.e., the chest X-ray dataset)~\cite{sowrirajan2021moco, azizi2021big}. 

We thus propose a hypothesis that semantic unawareness in image augmentation degrades the quality of contrastive learning and accounts for the above performance gap. We design controlled experiments on the publicly available Knee Osteoarthritis Severity Grading Dataset~(OAI) of X-ray images in Section~\ref{sec_best_recipe}. We show that, for the exemplar augmentation operations of \textit{cutout} and \textit{random resized crop}, the sweet spot (i.e., the optimal augmentation setting) is weaker in strength for medical images than the commonly used settings in natural images.
And the performance starts to drop earlier if the augmentation strength exceeds the sweet spot and becomes too much aggressive for the pre-training. 
To this end, we conclude that the augmentation recipe currently popularly used in natural images may not be suitable for medical images.

Further, this paper aims to propose attention-based augmentation to preserve image semantics that is critical to downstream diagnosis tasks. We argue that optimal contrastive learning should follow the human visual system, i.e., not to lose the salient parts in the images that provide critical visual cues to clinical diagnoses.
We thus design FocusContrast, an augmentation strategy guided by radiologists' vision, to replace the current handcrafted strategies for random image augmentation. The expert visual attention is collected by the eye-tracker attached to the monitor. The gaze collection is \textbf{seamless} and \textbf{effortless} in the daily clinical workflow~\cite{wang2022follow}.
Supervised by radiologists' gaze, we first learn to predict visual attention (or the gaze map) when a human expert reads a medical image. 
Based on the predicted gaze map, FocusContrast can filter out improper data augmentation, i.e., the transformation that possibly corrupts the salient part of the image.

The implementation of FocusContrast is straightforward.
Specifically, the learned augmentations can be summarized into two categories. 
\begin{itemize}
    \item We propose the \textit{focus cutout} and \textit{focus crop} to only remove the non-salient area of the image selectively. In this way, we can preserve the diseased abnormalities in positive pairs, even though they are small and easy to lose in previous random augmentation. 
    \item By learning from the radiologists who have the expertise in reading medical images efficiently, the \textit{focus mask} suppresses these non-salient areas. It makes the network easier to distinguish the difference between the negative pairs. 
\end{itemize}

FocusContrast is plug-and-play and a framework-agnostic module for current contrastive learning frameworks.
Experimental results show that the proposed semantic-aware augmentation consistently outperforms the conventional way of random augmentation or using hand-tuned hyper-parameters in MoCo~\cite{he2020momentum}, SimCLR~\cite{chen2020simple}, and BYOL~\cite{grill2020bootstrap}.

This paper is organized in the following manner.
In Section~\ref{sec:background}, we briefly cover the literature on contrastive learning and focus on the issue of positive pairs particularly. We also introduce the relevant attempts to use eye-tracking for radiology.
In Section~\ref{sec:optimal_view}, we design and analyze controlled experiments to illustrate why the current augmentation strategy may not be optimal for medical images. 
In Section~\ref{sec:learn_views}, we present FocusContrast, which to our knowledge is among the earliest works to utilize human visual attention to guide self-supervised learning. 
We will conclude this paper with more discussion in Section \ref{conclusion}.
Our code is publicly available\footnote{\href{https://github.com/JamesQFreeman/MICEYE}{https://github.com/JamesQFreeman/MICEYE}.}.

\section{Background}
\label{sec:background}
In this section, we first review self-supervised learning and contrastive learning for medical images. Then, we introduce the related works about radiologists' gaze and visual attention.

\subsection{Contrastive Learning and its Usage to Medical Images}

Contrastive learning has emerged as the front-runner for self-supervised learning, which has attained superior performance on downstream tasks. 
These pre-trained networks appended with linear probing can surpass the same fully-supervised networks, while only the appended linear layer is trainable~\cite{tomasev2022pushing}. Researchers also use linear probing to inspect the quality of the learned representations, i.e., by fixing the extracted features of the pre-trained models and evaluating their performance in the subsequent tasks. 
In general, large-scale contrastive pre-training with linear probing has become popular due to generalizability to many scenarios and robustness against overfitting~\cite{radford2021learning}.

There are several attempts to utilize contrastive pre-training in medical image analyses. Sowrirajan et al.~\cite{sowrirajan2021moco} proposed MoCo-CXR, which was an adaptation of the contrastive learning method of MoCo, to produce the models with better representations and initializations for abnormality detection in chest X-rays.
Azizi et al.~\cite{azizi2021big} utilized multiple images per patient to construct more informative positive pairs for self-supervised learning. With their proposed self-supervision paradigm, the network was able to outperform the supervised baseline.
Zhou et al.~\cite{zhou2020comparing} presented comparing-to-learn (C2L), providing pre-trained 2D deep models for radiograph-related tasks from unannotated data. 
All these works have adopted a conventional augmentation strategy similar to SimCLR~\cite{chen2020simple}. 

In the early days of contrastive learning, high-quality representation requires a large number of negative pairs in a batch (e.g. 4096). SimCLR~\cite{chen2020simple} has demonstrated that contrastive learning benefits from larger batch size, which allows for more negative pairs.
In order to increase the batch size, Momentum Contrast (MoCo)~\cite{he2020momentum} updates a momentum memory bank of negative representations to get rid of GPU memory restriction.
However, more recently, negative pairs are shown to be less necessary for representation learning. Several works show that the number of negative pairs may have a limited influence on the representation quality when the learning framework is designed in a sophisticated way. 
For example, BYOL~\cite{grill2020bootstrap} uses two Siamese networks and an extra non-linear transform, which enables contrastive learning with fewer negative samples in a single batch. 
DINO~\cite{caron2021emerging} has further extended this idea and utilized the transformer architecture, with a small batch size of 8.
As a summary, recent researches tend to pay more attention to positive pairs in contrastive learning. 
And in this paper, we will also focus on their roles in terms of the quality of the learned representations.


\subsection{Positive Pair Augmentation in Contrastive Learning}
One of the major design choices in contrastive learning is how to generate positive pairs. 
An intuitive approach is to create different views from the same sample using augmentation.
And most contrastive learning frameworks apply augmentation by adapting from the conventional way in supervised learning~\cite{chen2020simple,he2020momentum,grill2020bootstrap,cubuk2018autoaugment,cubuk2020randaugment}. 
Chen et al.~\cite{chen2020simple} comprehensively studied the augmentation operations. They found that more aggressive augmentation was usually needed in contrastive pre-training than supervised learning.
Tian et al.~\cite{tian2020what} proposed an InfoMin principle to catch a sweet spot of mutual information between augmented views, which indicated the maximal learning performance.

There are also recent works for semantic-aware augmentation.
Selvaraju et al.~\cite{selvaraju2021casting} proposed CAST to use unsupervised saliency maps to sample the cropping.
Peng et al.~\cite{peng2022crafting} proposed ContrastiveCrop for the augmentation of semantic-aware cropping. However, as we will demonstrate in the experiment section, medical images have very different semantic patterns compared to natural images, implying that those methods may not be suitable.
\subsection{Eye-tracking in Radiology}

Visual attention is a useful tool to understand and interpret radiologists' reasoning and clinical decision. 
In 1981, Carmody et al.~\cite{carmody1981finding} published one of the first eye-tracking studies in the field of radiology, where they studied the detection of lung nodules in chest X-ray films. Four radiologists participated and their eye movements were recorded using special glasses based on the corneal reflection technique. The participating radiologists were instructed to press a key when they found a nodule in the X-ray film. The study found that the false-negative errors in reading the X-ray data could be impacted by the eye-scanning strategies used by individual radiologists.

Eye-tracking studies are also conducted on other specialties such as mammography. Kundel et al.~\cite{kundel2008using} gathered eye-tracking data and found that 57\% of cancer lesions were located within the first second of viewing. Voisin et al.~\cite{voisin2013investigating} investigated the association between gaze patterns and diagnostic performance for lesion detection in mammograms. It is found that gaze fixations are highly correlated with radiologists’ diagnostic errors. 

There are also studies that focus on volumetric CT and MRI images. Bertram et al.~\cite{bertram2016eye} investigated the image markers of visual expertise using abdominal CT. 
The eye-tracking data shows that specialists react with longer fixations and shorter saccades when encountering the presence of lesions. 
Mallett et al.~\cite{mallett2014tracking} focused on CT colonography videos, which were interpreted by 27 experienced radiologists and 38 inexperienced radiologists. The eye-tracking data indicates that experienced readers have higher rates of polyp identification than inexperienced readers as evidenced by multiple pursuits when examining polyps. 
Stember et al.~\cite{stember2020integrating} used eye-tracking data to label brain tumors in MRI scans. 
More related studies are surveyed in \cite{brunye2019review, wu2019eye}.

\begin{figure*}[]
    \centering
    \includegraphics[width=0.99\textwidth]{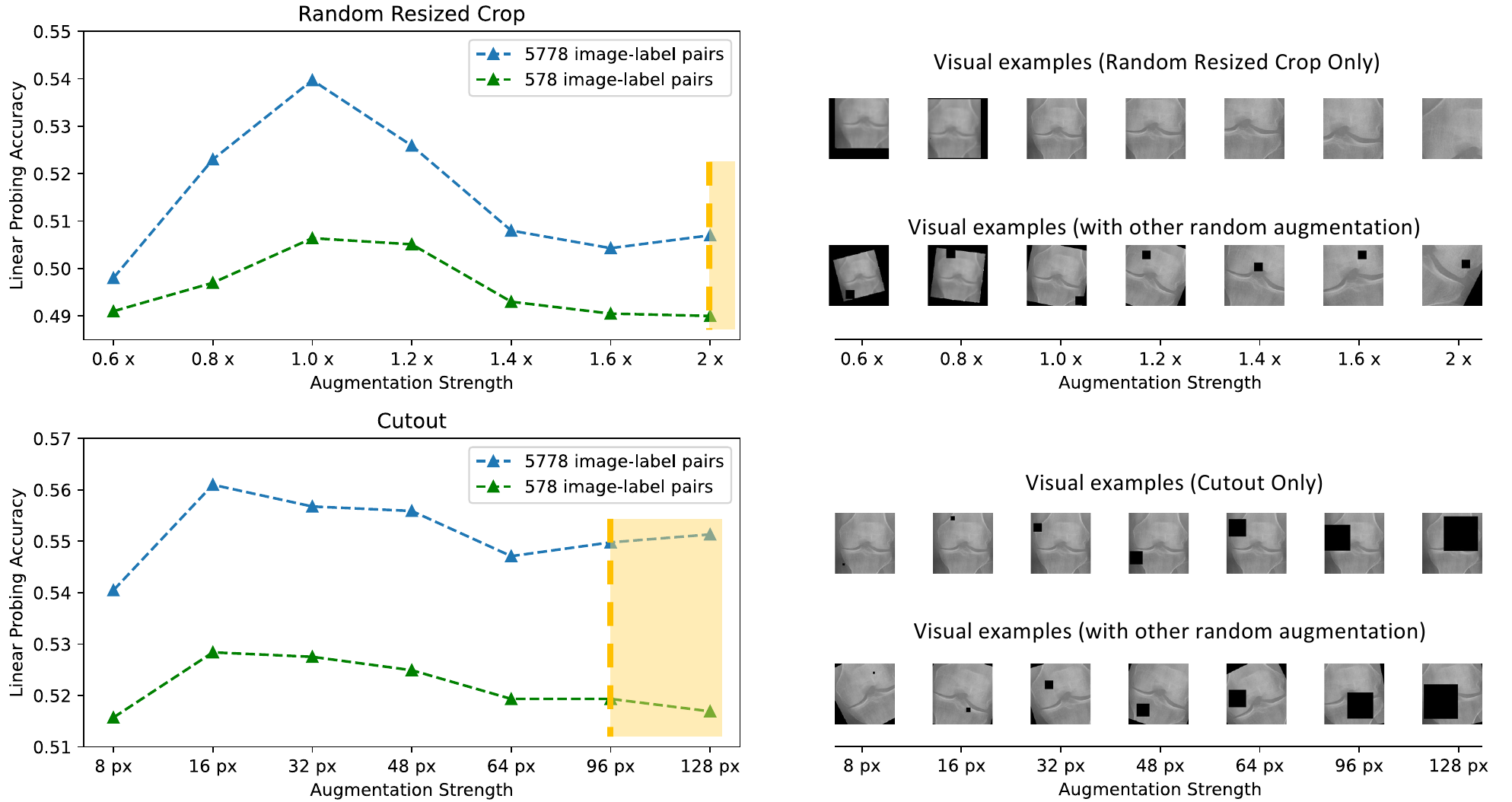}
    \caption{Relationship between the linear-probing classification performance and the augmentation strength in knee X-ray OA diagnosis. Random resized crop and random cutout are investigated. In the left panel, we report the linear-probing accuracy curves for 10\% label fraction (578 image-label pairs) and 100\% label fraction (5778 image-label pairs), with varying augmentation strength. The yellow areas in the curves denote the common augmentation settings in the natural image. In the right panel, we offer visual examples corresponding to different augmentation strength. }
    \label{fig:optimal-aug}
\end{figure*}

\section{What Are the Optimal Contrastive Views for Medical Image?}
\label{sec:optimal_view}
In this section, we first recall the InfoMin principle \cite{tian2020what}, which offers an explanation of why different tasks need different augmentation. Then, we illustrate this issue by using knee X-rays as an example, which is a common diagnostic examination in clinical practice.

\subsection{The Sweet Spot for Contrastive Augmentation}

When adopting contrastive learning for pre-training, two views $v_1$ and $v_2$ augmented from an image $x$ are often used to get $z_1$ and $z_2$. 
The objective of contrastive self-supervised learning maximizes the lower bound on the mutual information (MI) estimate between the two views, or $I(v_1; v_2)$ \cite{poole2019variational,soatto2014visual}.
However, in a downstream task such as classification for disease diagnosis, the objective is very different from the above in contrastive learning. 
For the classification task to predict the label $y$, the optimal representation $z^*$ can be encoded from the input image $x$.
Ideally, a model built with $z^*$ has all the information necessary to predict $y$ as accurately as accessing $x$ directly, or $I(z^*;y)=I(x;y)$. 
That is, no information loss occurs in mapping $x$ to $z$ concerning the task of predicting the label $y$ under the minimal sufficient statistic assumption.

The augmented views are critical to encoding the representations, which focus on the shared information between the views in contrastive learning \cite{oord2018representation}. If the shared information between a pair of weakly augmented images is too much, or $I(v_1;v_2)>I(z^*;y)$, then the captured representations cannot discard enough nuisance information from the inputs \cite{tian2020what}. That is, the code of $z$ carries many task-independent noises, which affect performance in the subsequent task due to reduced signal-to-noise ratio of $z$. On the other hand, if the key information to solve the classification task is missing, e.g., $I(v_1;v_2)<I(z^*;y)$, due to excessively strong augmentation, the code of $z$ also degrades the classification performance. As a result, Tian et al. \cite{tian2020what} defined the sweet spot in terms of the downstream performance where the useful information between the views is properly preserved following
\begin{equation}
    \label{eq-1}
    I(v_1;v_2)=I(z^*;y).
\end{equation}

We further argue that, for many classification tasks in medical images, the measure of $I(z^*;y)$ can be high whereas the diseased abnormalities are often tiny. Note that $z$ is usually computed as a spatial integration of localized features, e.g., by pooling from the convolutional network outputs. Thus, the entropy of $z$ conditioned on the label $y$ is small, since $z$ is dependent on restricted spatial locations that are abnormal and thus associated with the disease diagnosis. Meanwhile, as the conditional entropy of $z$ upon $y$ is an inverse to $I(z^*;y)$, we conclude that $I(z^*;y)$ is high particularly for the medical images such as in Fig. \ref{fig:hypothesis}. 

Therefore, it leads to our assumption that augmentation strength should be relatively weaker corresponding to the sweet spot for medical images than the common settings in natural images. Referring to equation~\eqref{eq-1}, a too strong augmentation may reduce mutual information between $v_1$ and $v_2$ significantly, which induces information loss to the learned representations and affects the downstream classification.

To make sure that $I(v_1;v_2)$ is kept high throughout image augmentation, besides controlling the augmentation strength, it is necessary to preserve the region of the diseased abnormality in the image. In this way, $I(z_1;z_2)$ should be close to $I(v_1;v_2)$, as the mapping from the input image to its latent representation may suffer little from the information loss. In this way, one can guarantee the key information to pass to the downstream classification task. Meanwhile, the redundant features that are influence subsequent task are removed from learned representations. 

\subsection{Augmentation Recipes for Knee X-rays}
\label{sec_best_recipe}

We investigate two widely-adopted augmentations to illustrate the shifting sweet spots by using knee X-rays as the example. We pay special attention to two operations of \textit{random resized crop} (to crop a random portion of the image and then resize the cropped image to the original size) and \textit{random cutout} (to randomly remove a portion of the image)~\cite{devries2017improved}. The two operations could affect the learned representation, since they could easily remove the semantic cues in the images when the diseased abnormalities are small. We first briefly introduce our settings and then present the experimental results.

\textbf{Dataset and follow-up task.} The knee X-ray images are acquired from the OAI repository, which is made publicly available by \cite{chen2019fully}. 
The dataset used in this paper includes 5778 training images and 1656 testing images. 
In the contrastive pre-training stage, all the training images are used. 
Then, to evaluate the learned representations of the contrastive pre-training, the follow-up linear probing task is trained with ``10\% label fraction'' and `` 100\% label fraction" setting. That is, we use 578 and 5778 training images and their class labels to train the classifier. Then, we apply the trained classifier to the test images, and yield their labels prediction in five classes. Specifically, the knee OA severity is measured by the Kellgren-Lawrence (KL) grading system~\cite{kellgren1957radiological}, which is a 5-point semi-quantitative progressive ordinal scale ranging from Grade 0 (normal) to 4 (high severity).
We use the KL scores as target of the classification task.

\textbf{Contrastive learning framework.} We choose BYOL as the contrastive learning framework here. Specifically, We train at batch size 128 for 100 epochs on Nvidia RTX Titan. Furthermore, we use linear warm-up for first 10 epochs, and decay the learning rate with cosine decay schedule. Regarding the loss, we use InfoNCE optimized by LARS optimizer with the learning rate of 0.2 and weight decay of $1.5e^{-6}$. We use ResNet-50~\cite{he2016deep} as our backbone. 
Our experimental settings may not ensure the best possible performance (e.g., bigger backbone, longer training and larger batch size that can further boost the performance), but they do allow us to validate our hypothesis. 

Although we are interested in the augmentations of \textit{random resized crop} and \textit{random cutout}, we also deploy other popular image augmentations for contrastive learning. 
The reason is that, if only \textit{random resized crop} or \textit{random cutout} is used, the strictly controlled augmentation setting would be too far away from the current real case. 
To this end, in our experiments, we resize all images to 224$\times$224, followed by random flip, random color distortion, random rotation (30 degrees), random cutout (32px$\times$32px) and random resized crop (1.2$\times$ zoom-in) by default. 
When we investigate the impact of \textit{random resized crop} for example, we will change it from the default to our specific setting, while keeping other augmentation settings unchanged.


\textbf{Results for \textit{Random Resized Crop}.} In Fig. \ref{fig:optimal-aug} Row 1, we report the results when \textit{random resized crop} varies its strength. 
Specifically, we test 0.6$\times$, 0.8$\times$, 1$\times$, 1.2$\times$, 1.4$\times$, 1.6$\times$, and 2$\times$ for random resized crop. 
Given the 1.4$\times$ setting as an example, the 224$\times$224 image is first resized to 313$\times$313, and then cropped back to a 224$\times$224 output with a randomly chosen center location. 
The accuracy curve for the ``578 image-label pairs'' shows a peak around the 1.0$\times$-1.2$\times$ settings. 
And accuracy starts to drop for stronger augmentation such as 1.4$\times$. By comparison, the sweet spot is usually set to 2$\times$ for natural image in contrastive learning, which is significantly stronger than the optimal augmentation strength acquired for knee X-rays in our experiment. 
As a reference, we also provide ``5778 image-label pairs'' curve in the figure. 
And the same trend can be observed, implying that adding more supervision to the classification task cannot easily compensate for the bad representations caused by improper augmentation to contrastive pre-training.

In the top-right of the figure, we show examples of the augmented images with varying strength of \textit{random resized crop}. One may easily notice that the images change their appearances significantly when only \textit{random resized crop} is enabled and its strength is increasing. While other augmentations are also turned on as in our real experiments, the effect of \textit{random resized crop} is still clear visually. For example, at 2$\times$, the OA-related region of the femur-tibia interface is cropped out, implying that the augmentation is excessively strong. Meanwhile, one may notice that 0.6$\times$-0.8$\times$ settings can shrink the original images by padding zero-valued background. In this way, the very weak augmentation induces much distortion in the augmented images, which also echos the relatively low classification performance achieved by linear probing.

\textbf{Results for \textit{Random Cutout}.} In Fig. \ref{fig:optimal-aug} Row 2, we report the linear-probing accuracy when \textit{random cutout} varies its strength. Specifically, we test 8px, 16px, 32px, 48px, 64px, 96px, 128px, as the value indicates the size of the cutout region. 
One can notice that \textit{random cutout} larger than 64px (i.e., larger than 8.2\% of the whole image) will degrade the representations learned from the Siamese contrastive learning. In comparison, \textit{random cutout} in the latest works like SwinTransformer~\cite{liu2021swin} and ConvMixer~\cite{trockman2022patches} can be tuned up to 1/3 of the field-of-views for natural images, which is much larger than our finding in the knee X-ray images.

In the right half of the figure, we show examples of augmented images with varying cutout sizes. One may easily observe that the knee joint, which is critical for diagnosis, can be masked by a large cutout area when the strength of \textit{random cutout only} is creasing. 
Despite turning on other augmentations like random resized crop (1.2$\times$ in our example) will increase the area of the diseased abnormality, \textit{random cutout} still masks out too much. 
For example, at 96px and 128px, the knee joint and cartilage are masked out, resulting in potentially incorrect positive pairs in contrastive training.

\textbf{Conclusion.} We conduct controlled experiments on two widely-used data augmentations of \textit{random resized crop} and \textit{random cutout} by testing different augmentation strength. Particularly for knee X-ray images, we empirically find the optimal augmentation hyper-parameters, which are (1$\times$,1.2$\times$) for \textit{random resized crop} and (16px, 48px) for \textit{random cutout}. 
Both settings are significantly weaker than those in natural images. 
The results demonstrate that aggressive augmentations noticeably degrade the representations from contrastive learning. This is in line with our expectation since (1) subtle lesions can be removed by aggressive random augmentation, and (2) the transformed image is out of distribution and thus cannot benefit the follow-up task. 

\begin{figure}[t]
    \centering
    \includegraphics[width=0.49\textwidth]{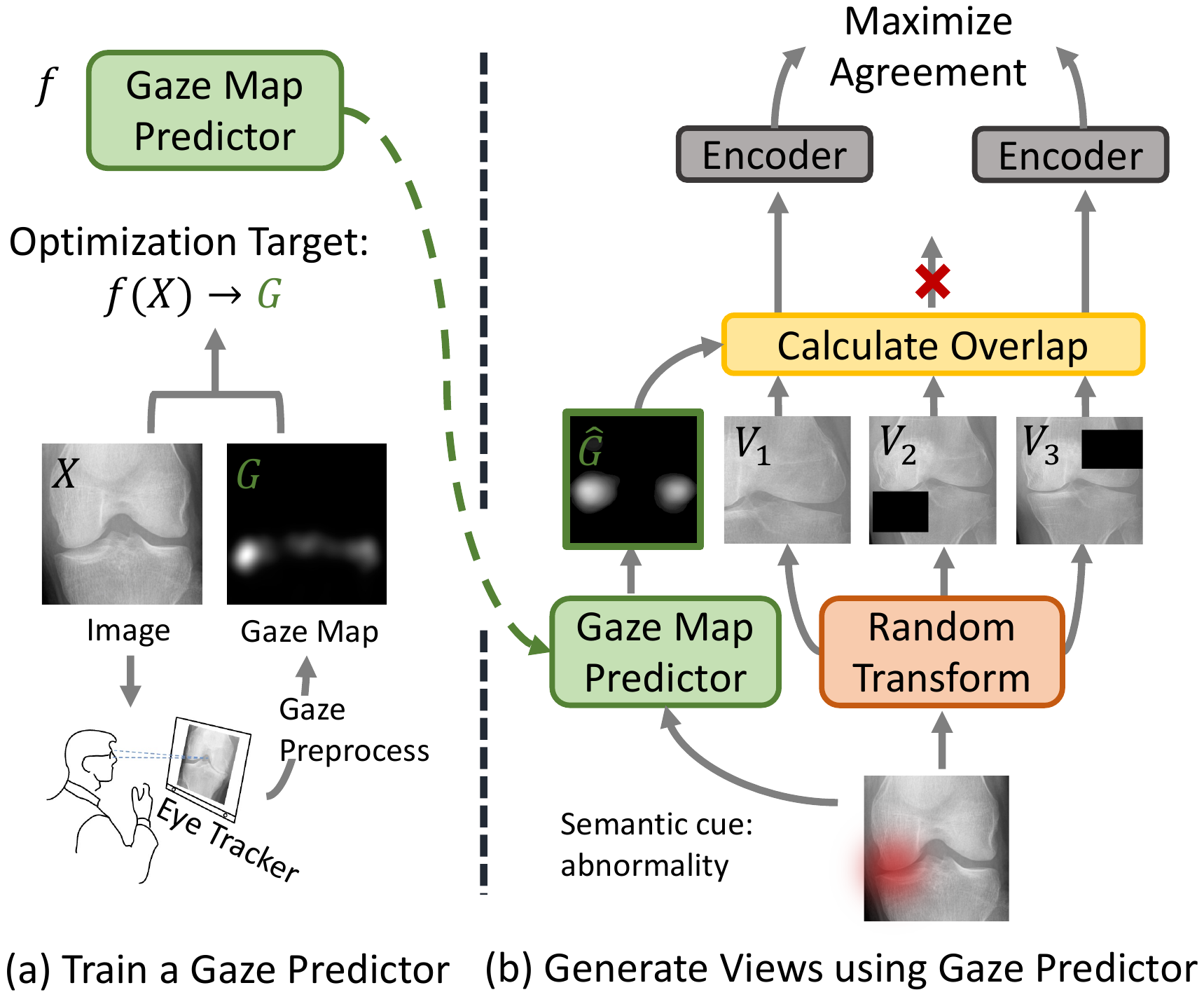}
    \caption{The overview of FocusContrast augmentation framework. In (a), a network called gaze map predictor is trained to predict the radiologist's gaze map. In (b), when an image is augmented into different views, the gaze map predictor helps distinguish whether the salient parts in the images are preserved or not. If the salient part is preserved, the augmented view will be fed to the encoder for contrastive learning; otherwise, it will be treated as a false-positive view and discarded.}
    \label{fig:pipeline}
\end{figure}

\section{Learning Views from Radiologists for Contrastive Learning}
\label{sec:learn_views}
In Section \ref{sec:optimal_view}, we demonstrate that it is not appropriate to simply borrow the contrastive augmentation transforms in natural images and then apply them to medical images. 
And we conduct extensive experiments on two popular augmentations by searching for their optimal hyper-parameters in brutal force.
However, an exhaustive search over all possible hyper-parameters is not always a good solution in that it is time-consuming and expensive. 

To solve this problem, we propose a straightforward yet efficient method to learn the semantic-aware augmentations from the visual attention of the radiologists for contrastive views. 
As shown in Fig. \ref{fig:pipeline},
we first collect hundreds of gazes from radiologists, and train a gaze map predictor. 
Then, we use the trained gaze map predictor to judge each augmented view and decide whether the view is qualified to feed into contrastive learning. 
In particular, the gaze map predictor tells which part in the (augmented) image contains salient semantics according to the learned radiologists' gaze. 
If the salient part is preserved in the augmented view, it will be used in contrastive learning; otherwise, it will be discarded.

In the next, we use the same images and task as the last section to compare our learnable FocusContrast with the brutal-force ones. 
We will first introduce how to collect the gaze data, followed by the training of the gaze map predictor. 
Then we will introduce the way to filter out the inappropriate views to eliminate false positives for contrastive learning. 
Finally, we will demonstrate that our method can be highly effective in contrastive representation learning.

\subsection{Gaze Collection and Post-processing}
\label{gaze collection and post-processing}
\begin{figure}[t]
    \centering
    \includegraphics[width=0.48\textwidth]{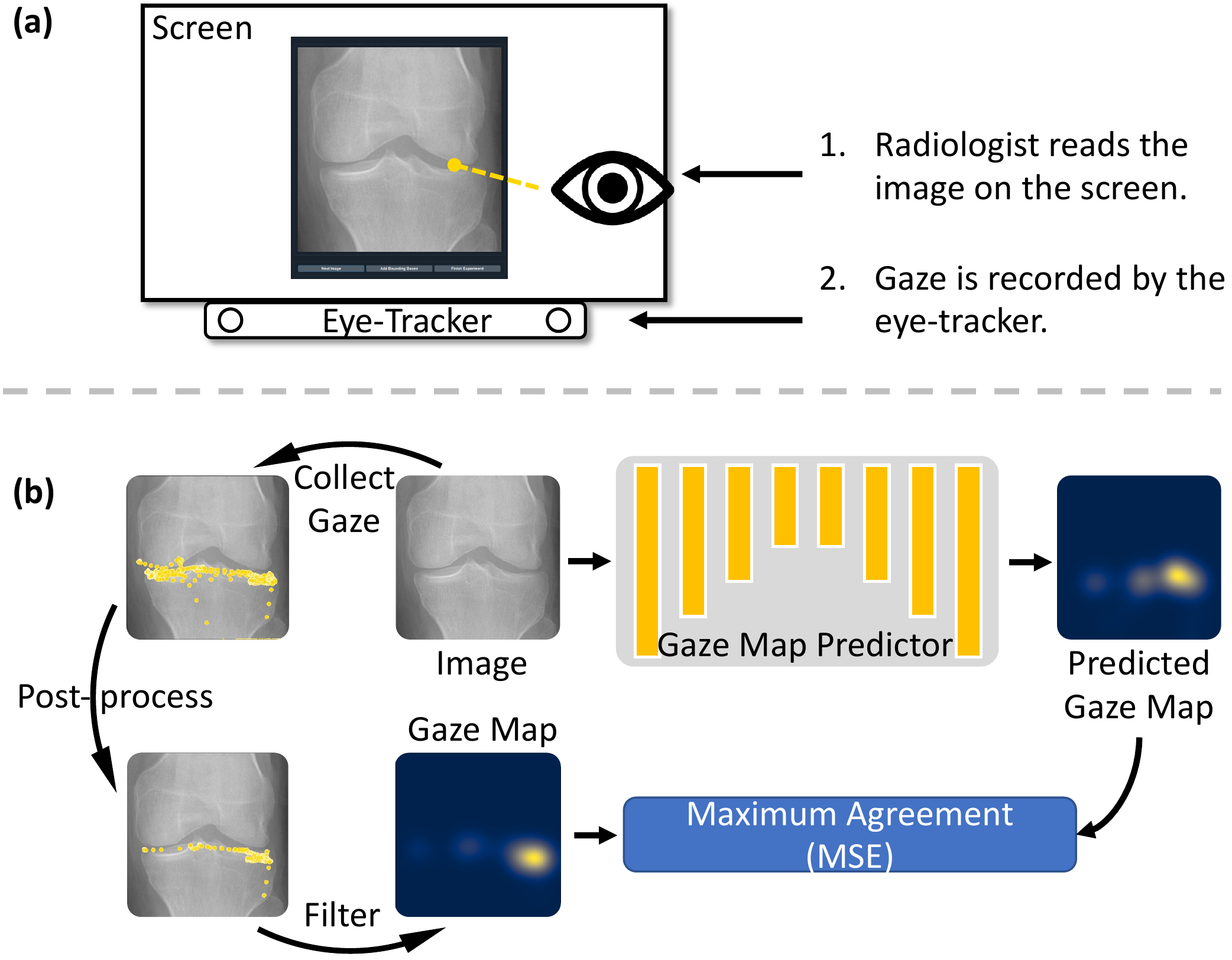}
    \caption{(a) Radiologists look at a knee X-ray image until they make a diagnosis decision. From the moment they are shown this image to the moment they type down their decision, their eye movement is recorded by the eye-tracker. (b) The eye movement is recorded discretely in coordinate-timestamp form. The gaze points are first processed to remove the distortion. Then a Gaussian kernel is applied to the gaze points, which generates the gaze map (or the saliency heatmap). The image and the gaze map are used as the ground-truth pair to train the gaze map predictor.}
    \label{fig:gaze-collect}
\end{figure}

\begin{figure*}[h]
    \centering
    \includegraphics[width=0.99\textwidth]{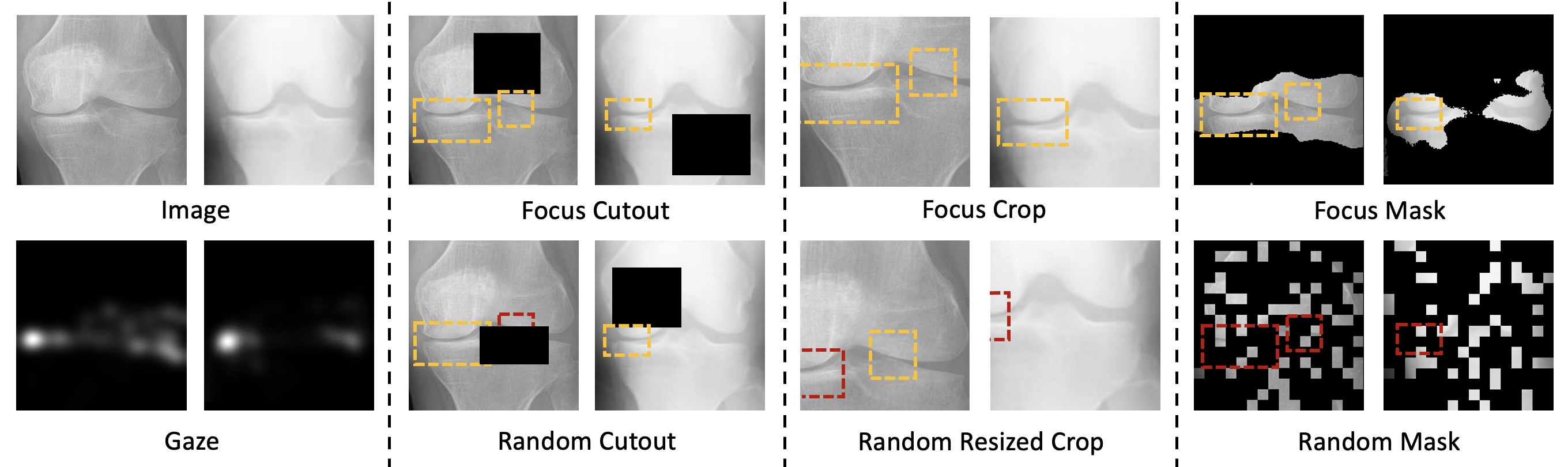}
    \caption{On the left, we demonstrate an image and its 
corresponding gaze map from a radiologist. On the right, the image is augmented to generate the positive pairs. We present the exemplar images after augmenting by FocusContrast (top row), and compare them with the conventional counterparts (bottom row). We use the orange boxes to indicate that the semantic cues (i.e., abnormal areas related to disease diagnosis) are preserved in the augmented images. Contrarily, the red boxes indicate the key areas that are missing or partially corrupted after conventional augmentation.}
    \label{fig:augmentation}
\end{figure*}
We collect the eye-tracking data with the Tobii 4C remote eye-tracker that records binocular gaze data at 90Hz. We implement customized data collection software in Python using the manufacturer-provided SDK. The software is made publicly available with this paper.  
Readers are seated in front of a 27-inch LCD screen, to simulate the clinical working condition. They can adjust their seat distances to the screen for comfort.
Our software logs the reader's on-screen gaze locations with the corresponding timestamps.

Our data collection paradigm is illustrated in Fig. \ref{fig:gaze-collect}(a). Note that we make the demonstration by reading knee X-ray images for OA assessment. 
First, radiologists log in their profiles of identity. And then we calibrate the eye-tracker using the standard 5-point calibration routine~\cite{karessli2017gaze} for the participating radiologists. 
Next, the radiologist follows a cycle of two steps, namely reading and diagnosis, to complete the diagnosis task on an X-ray image. 
During the reading step, an image is randomly drawn from our training dataset and shown to the radiologist, as the radiologist reads the image until they feel confident to reach a diagnosis decision. 
In the diagnosis step, the radiologist types in the decision by pressing the number keys on the keyboard, e.g., ``1-4” used for representing KL-Grades 1-4, and ``Enter"  for normal or Grade 0. We record eye movement during the entire cycle of reading and diagnosis steps. 
We have a rest of 2 minutes for every 20 images to reduce radiologists' fatigue. 
In this paper, we use 354 gaze tracks, containing 154 images of Grade 0, 55 images of Grade 1, 81 images of Grade 2, 40 images of Grade 3, and 24 images of Grade 4.

\subsection{Training of the Gaze Map Predictor}

The training procedure is demonstrated briefly in Fig. \ref{fig:gaze-collect}(b). As the figure shows, the collected gaze point sequence is filtered by a post-processing algorithm~\cite{wang2022follow} to remove the gaze points when the radiologist is looking at the GUI elements or other distractions. 
We further acquire the gaze map from the filtered gaze points per training image, by modeling them in the Gaussian mixtures. 
The Gaussian kernel is sized to 99, determined by the human para-central vision area~\cite{carrasco2011visual}, image size ($25\times 25 cm^2$) and distance from eyes to screen~($\approx50cm$)~\cite{wang2022follow}. 

The X-ray image and the gaze map are used as the input and output to supervise the training of a U-Net~\cite{ronneberger2015u}. The trained U-Net of gaze map predictor can predict where radiologists will pay attention when reading a knee X-ray image. 
The U-Net consists of 4 downsample encoders and 4 upsample decoders, and has the same size for the input and output. The network is trained using Adam~\cite{kingma2014adam} for 30 epochs and the weight of the last epoch is used in our experiment. Empirically, a minor mistake of the gaze map predictor will not influence the follow-up view-filtering process as we only need it to give us a rough location of where the radiologist would pay attention to.

\subsection{Generating Image Views using Gaze Map Predictor}
The intuition of our method is simple -- when radiologists think the specific area of the image is informative, these areas should not be lost during the augmentation to prevent potential information loss. 
Based on this idea, we propose three novel augmentations to generate the positive views using our trained gaze map predictor: 1) \textit{focus cutout}, 2) \textit{focus crop}, and 3) \textit{focus mask}. 
These three augmentations are designed based on the classical \textit{cutout}, \textit{random resized crop} and \textit{random mask}, respectively.
See Fig. \ref{fig:augmentation} for visual examples of how these augmentations are done.  

\textbf{Focus cutout.} Cutout is a commonly used augmentation technique that improves the robustness and overall performance of networks~\cite{devries2017improved, zhong2020random}. 
It randomly masks out square regions of the input, which could accidentally remove the important semantic information, e.g. the diseased abnormality. 
To prevent such loss, we propose \textit{focus cutout}, which only randomly masks out regions that radiologists may not pay attention to. 

The implementation of \textit{focus cutout} is detailed in Algorithm \ref{code}. 
Specifically, we predict the gaze map for an input image $x$. Follwing a random cutout, we get an augmented view $v_1$ as well as its corresponding gaze map. 
Then, we calculate the \textit{intersection} (i.e., IOU) between the non-zero gaze-map regions before and after random cutout. 
If the IOU scores for two augmented views ($v_1$ and $v_2$) are both above a threshold (e.g., 0.9), the two views are preserved as a valid positive pair. 

\algdef{SE}[DOWHILE]{Do}{doWhile}{\algorithmicdo}[1]{\algorithmicwhile\ #1}%
\algrenewcommand{\Return}{\State\algorithmicreturn~}
\begin{algorithm}
\caption{Example of focus cutout}\label{alg}
\begin{algorithmic}
\Require $x$ is an input image
\Require network for gaze map predictor
\State $gaze$ = Predictor($x$)
\Do\Comment{Same cutout for both views}
  \State $v_1$, $gaze_{v1}$ = random\_cutout([$x$, $gaze$])
  \State $v_2$, $gaze_{v2}$ = random\_cutout([$x$, $gaze$])
\doWhile{IOU($gaze_{v1}$,$gaze$)$>0.9$ \& IOU($gaze_{v2}$,$gaze$)$>0.9$}
\Return $v_1$, $v_2$\Comment{A positive pair after augmentation}
\end{algorithmic}
\label{code}
\end{algorithm}

\textbf{Focus crop.} In SimCLR, \textit{random resized crop} is performed for ``context aggregation" and ``neighboring view prediction". However, an aggressive resized crop ($\geq$1.4$\times$ in Fig. \ref{fig:optimal-aug}) may remove the critical abnormality. 
In Fig.~\ref{fig:augmentation}, \textit{random resized crop} also has lost the abnormality in the knee joint (indicated by red boxes). 
To address this problem, we propose \textit{focus crop}. Our implementation is similar to \textit{focus cutout} -- 
zoom-in resizing is only performed when it does not remove too much area of radiologists' interest.
In particular, the \textit{intersection} threshold is set to 0.8 for \textit{focus crop}.

\textbf{Focus mask.} The idea of using \textit{dropout} or \textit{random mask}~\cite{srivastava2014dropout} for augmentation is natural and applicable in self-supervised learning. However, while this simple idea has achieved success in natural language processing~\cite{gao2021simcse}, it is not common in computer vision until very recently~\cite{he2021masked}. In this paper, we propose \textit{focus mask}, which masks the area that the reader is not much paying attention to. 
A threshold is applied to the gaze map to mask out 80\% less-informative areas.

\subsection{Experimental settings}

We inherit the knee X-ray dataset and the KL-grading task from Section~\ref{sec_best_recipe}.
In our experiments, following \cite{chen2020simple,he2020momentum,grill2020bootstrap}, we use ResNet50 as our backbone. If not otherwise specified, our models here are trained for 200 epochs with batch size of 512.
We use MMSelfSup~\cite{mmselfsup2021} framework to pre-train our backbone and adopt the recommended training recipe. During pre-training, all 5778 images in the training set are used, which is also the same with Section \ref{sec:optimal_view}.

To evaluate the effectiveness of the pre-trained representations, we freeze the backbone model and train a linear classifier on top using the labeled data (denoted as \textit{linear-probe} in our experimental results). In addition, we can unfreeze all layers and fine-tune the entire model end-to-end, to compare transferability on the overall performance (denoted as \textit{fine-tune}).
Both the linear-probing and the fine-tuning schemes under comparison are trained for 100 epochs with batch size of 128.  

We have validated five different sizes of the classification training sets: 1\% label fraction (55 images evenly across 5 classes contributing their labels to train the classification task), 5\% label fraction (288 images), 10\% label fraction (577 images), 20\% label fraction (1154 images) and 100\% label fraction (5778 images). 
All evaluation is conducted on 1656 testing images, and the best classification accuracy (ACC) is reported with the mean absolute error (MAE) of the predicted grades at the same epoch. 

\subsection{Experimental results}

\textbf{Comparisons to other attention prediction models.}
In this research, we use gaze to supervise the training for visual attention prediction. There are many different ways to predict human visual attention. However, professional radiologists have very different visual attention compared to non-professionals. Therefore, the attention prediction models designed for the general population and common visual tasks cannot function well in medical images. 
To verify the above, in this experiment, we compare our gaze map predictor with the commonly adopted methods of SpectralResidual (SR)~\cite{hou2007saliency}, Montabone et al.~\cite{montabone2010human}, and Pyramid Feature Attention Network (PFAN)~\cite{zhao2019pyramid}. Note that PFAN is regarded as a state-of-the-art saliency prediction algorithm trained on natural images, ranking top on PASCAL-S and DUT-OMRON~\cite{zhao2019pyramid}.

\begin{figure}[h]
    \centering
    \includegraphics[width=0.48\textwidth]{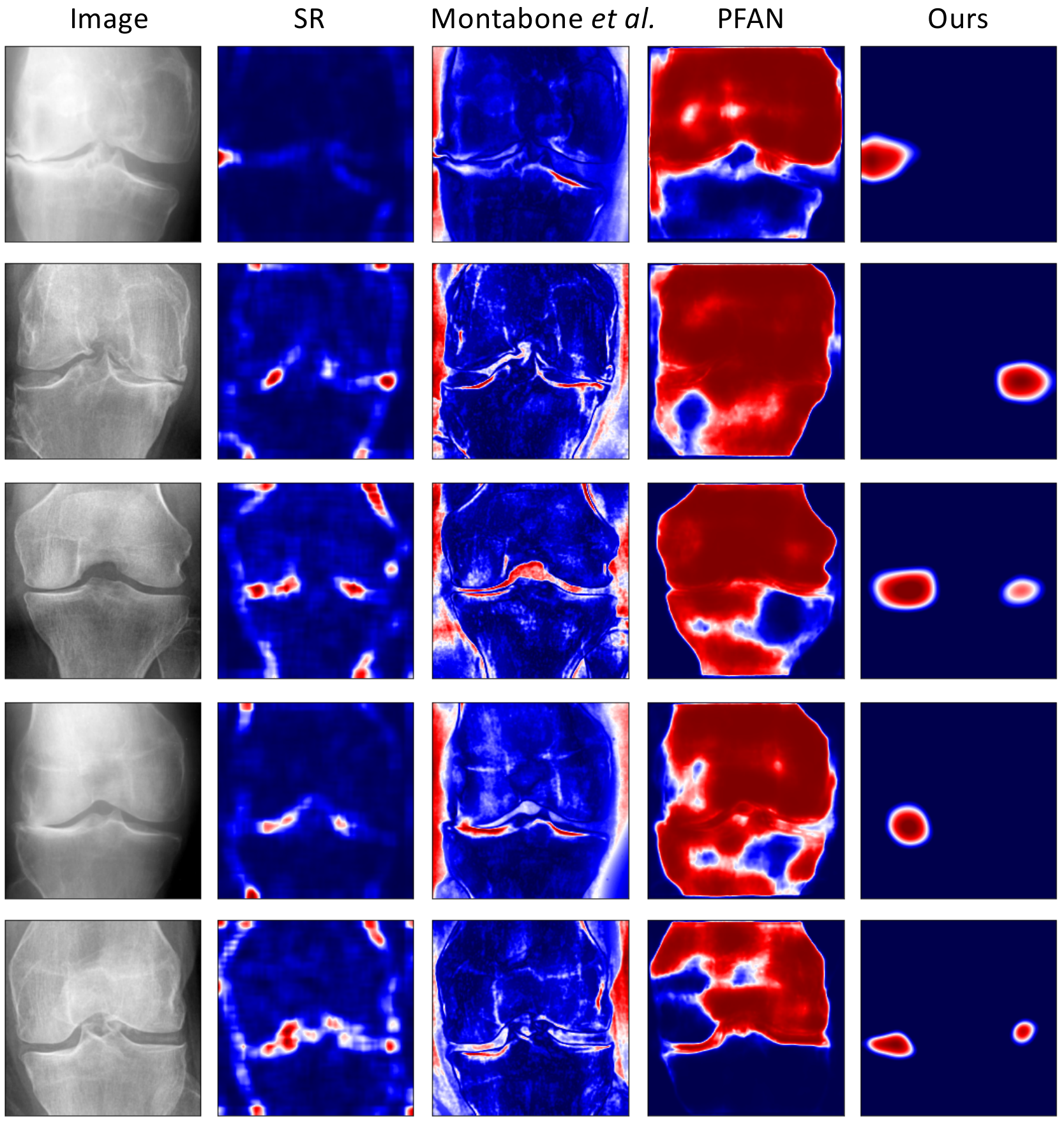}
    \caption{Comparison of different attention prediction models on knee X-ray images. Red denotes high attention while blue denotes low attention. Our model focuses more compactly on the diagnosis-related areas in the images.}
    \label{fig:different_saliency}
\end{figure}

The results are presented in Fig. \ref{fig:different_saliency}. Our attention prediction model trained on the radiologists' gaze easily outperforms other models, which are designed for the general-purpose human visual system. This is in line with our early finding, as radiologists are trained to seek the areas with high disease-related risks in an image and focus on them more \cite{wang2022follow}. 

In contrast, the general-purpose human visual system tends to look at highly informative areas (e.g., with large intensity changes). And the compared methods largely follow or mimic the human visual system as reflected by their predicted attention maps. 
For example, SR~\cite{hou2007saliency} summarizes from natural image statistics that 
intensity-frequency has a log-log relationship on average. Then, the area out of this log-log distribution leads to a high response in attention. On the other hand, PFAN uses a learning-based approach, training on paired natural images and eye-tracking data~\cite{zhao2019pyramid}. The prediction, however, cannot localize the critical areas effectively.

\textbf{Comparisons to handcrafted augmentations.}
In this section, we compare the learning-based FocusContrast with the handcrafted augmentations, which are widely used for contrastive pre-training.
We also compare with the optimal augmentation hyper-parameters searched exhaustively (i.e., 1.2$\times$ for random resized crop and 48px for cutout, see Section~\ref{sec_best_recipe}). The three settings are denoted by the superscripts \textit{Learned}, \textit{Default}, and \textit{Searched} in Table \ref{LearntVsDesigned}, respectively. 
Note that we use BYOL as the common contrastive learning framework for the comparisons here. 

In addition, there are three baselines to compare in the table, including: a randomly initialized ResNet-50 that is trained in the end-to-end manner (denoted as \textit{from-scratch}); a backbone fixed after ImageNet-based pre-training and a linear-probing classifier (denoted as \textit{IN linear}); a network pre-trained by ImageNet and then fine-tuned end-to-end (denoted as \textit{IN fine-tune}).
To investigate the training efficiency when utilizing classification labels, we also test our models on different label fractions of the training data. 

\begin{figure}[h]
    \centering
    \includegraphics[width=0.48\textwidth]{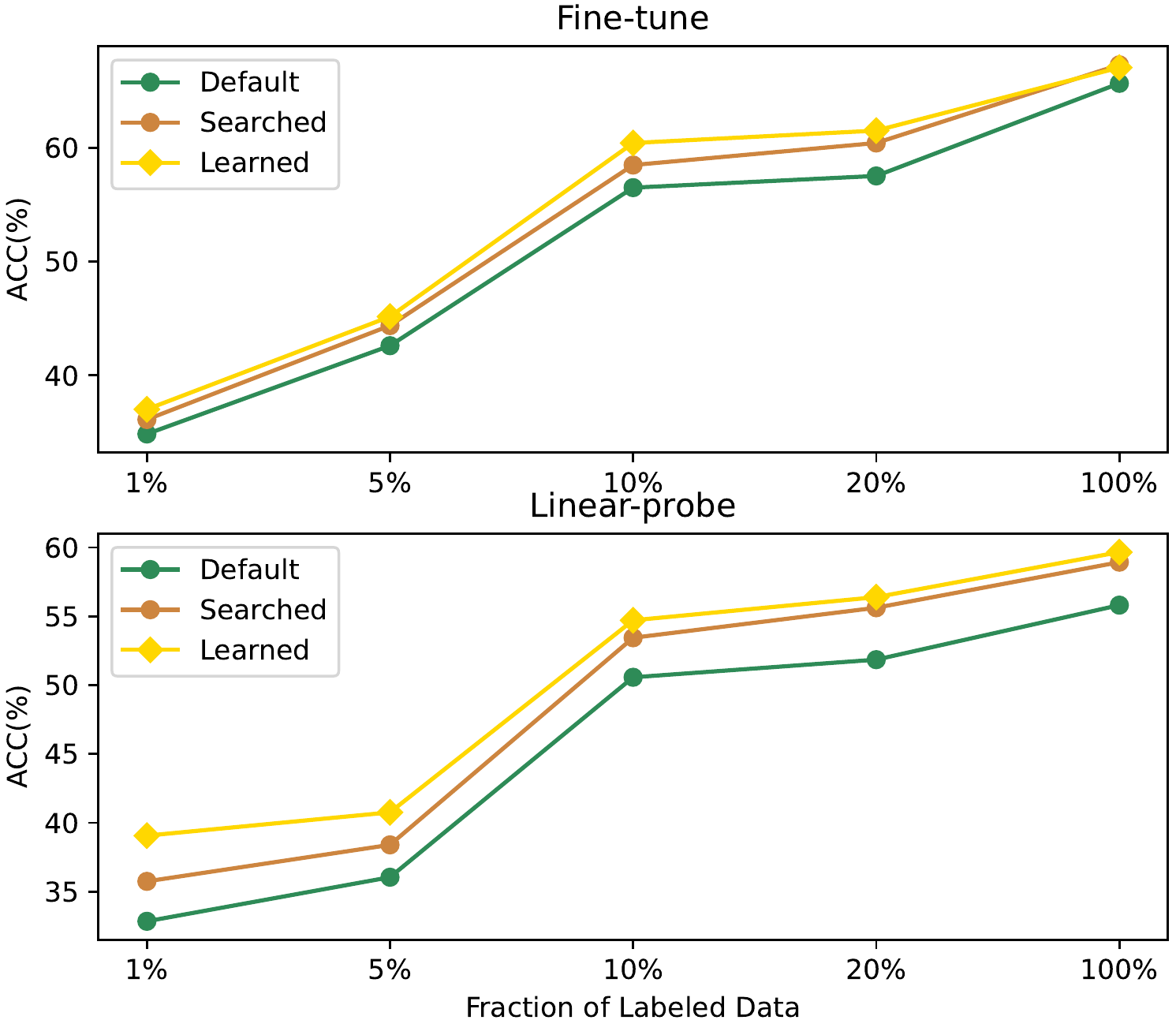}
    \caption{Classification accuracy using different augmentation strategies.}
    \label{fig:result_plot}
\end{figure}

There are several observations to derive from Table~\ref{LearntVsDesigned} and Fig.~\ref{fig:result_plot}.
\begin{itemize}
    \item The pre-training procedure, especially contrastive learning, improves the performance of the final classification task effectively. 
    In the first row of Table~\ref{LearntVsDesigned}, the accuracy of the network trained \textit{from-scratch} is significantly lower than other settings. Specifically, the ACC gap is +5.67\% between \textit{from-scratch} and \textit{IN fine-tune} when 100\% label fraction is available; the gap becomes even larger (+10.85\% ACC in average) when the label fraction is reduced to below 20\%.
    Compared to pre-training upon ImageNet, contrastive learning constantly improves the learned representations and the model's transferability over different label fractions of the training data.
    These observations are in line with other recent literatures~\cite{azizi2021big}.

    \item FocusContrast outperforms default contrastive learning in linear-probing performance. FocusContrast (denoted as Learned, yellow line in Fig.~\ref{fig:result_plot} bottom panel) constantly outperforms the widely adopted default random augmentations (denoted as Default, green line in Fig.~\ref{fig:result_plot} bottom panel) by a large margin, averaging in a +4.90\% ACC gap. 
    These results prove that FocusContrast helps to learn a higher quality representation than the default random augmentation does. 
    \item Compared to default augmentation, FocusContrast is more label-efficient to the subsequent task. For example, linear-probe$^{Learned}$ at 1\% label fraction (ACC: 39.07\%) performs even better than linear-probe$^{Default}$ at the 5\% label fraction (ACC: 36.05\%). And linear-probe$^{Learned}$ at 20\% label fraction (ACC: 56.39\%) performs better than linear-probe$^{Default}$ at 100\% label fraction (ACC: 55.81\%).
    \item As for transferability, FocusContrast (denoted as fine-tune$^{Learned}$, Row 6) still introduces considerable improvement compared to the widely adopted network parameters pre-trained from ImageNet (denoted as IN fine-tune, Row 2) and contrastive pre-training by default augmentations (denoted as fine-tune$^{Default}$, Row 4). 
    \item FocusContrast has a similar performance as the exhaustively searched optimal augmentation setting. 
    On the lower panel of Fig.~\ref{fig:result_plot}, FocusContrast (yellow line) has a larger linear-probing performance lead (compared to searched augmentation, orange line) when fewer labeled data is available. On the upper panel of the fine-tune performance, the performance gap is smaller.
\end{itemize}

In summary, FocusContrast, being learning-based, outperforms other widely-adopted approaches (ImageNet-based pre-training and contrastive learning with conventional augmentations).
It can even have a slightly better overall performance than the exhaustively searched (and thus optimal) augmentation setting. 
In that collecting gaze is seamless and effortless in our configuration while searching hyper-parameters is time-consuming and computationally intense, FocusContrast can be practical and useful to develop many future computer-aided diagnosis systems.


\begin{table*}[]
\centering
\caption{OAI KL-Grade prediction performance of models trained with different number of labels. ResNet-50 is used as the backbone.}
\begin{tabular}{lcccccccccc}
\hline
                          & \multicolumn{10}{c}{Label fraction}                                                                                                 \\
Method                    & \multicolumn{2}{c}{1\%} & \multicolumn{2}{c}{5\%} & \multicolumn{2}{c}{10\%} & \multicolumn{2}{c}{20\%} & \multicolumn{2}{c}{100\%} \\
                          & ACC        & MAE        & ACC        & MAE        & ACC         & MAE        & ACC         & MAE        & ACC         & MAE         \\ \hline
from-scratch              & 26.03      & 1.370      & 37.62      & 1.216      & 39.79       & 1.163      & 39.25       & 1.189      & 58.10       & 0.614       \\
IN fine-tune              & 33.94      & 1.028      & 38.95      & 1.007      & 55.19       & 0.624      & 58.03       & 0.619      & 63.77       & 0.502       \\
IN linear                 & 30.43      & 1.272      & 33.88      & 1.002      & 43.12       & 0.986      & 44.38       & 0.938      & 49.40       & 0.790       \\ \hline
fine-tune$^{Default}$     & 34.84      & 0.980      & 42.61      & 0.958      & 56.52       & 0.606      & 57.55       & 0.624      & 65.70       & 0.445       \\
fine-tune$^{Searched}$    & 36.11      & 0.974      & 44.38      & 0.874      & 58.51       & 0.579      & 60.45       & 0.565      & \textbf{67.33}       & \textbf{0.429}       \\
fine-tune$^{Learned}$     & \textbf{37.01}      & \textbf{0.930}      & \textbf{45.17}      & \textbf{0.908}      & \textbf{60.45}       & \textbf{0.559}      & \textbf{61.54}       & \textbf{0.557}      & 67.09       & 0.434       \\ \hline
linear-probe$^{Default}$  & 32.85      & 1.053      & 36.05      & 1.097      & 50.57       & 0.755      & 51.85       & 0.763      & 55.81       & 0.685       \\
linear-probe$^{Searched}$ & 35.75      & 1.018      & 38.40      & 1.026      & 53.44       & 0.697      & 55.62       & 0.651      & 58.93       & 0.586       \\
linear-probe$^{Learned}$  & \textbf{39.07}      & \textbf{0.961}      & \textbf{40.76}      & \textbf{0.872}      & \textbf{54.71}       & 0.\textbf{658}      & \textbf{56.39}       & \textbf{0.649}      & \textbf{59.66}       & \textbf{0.583}       \\ \hline
\end{tabular}
\label{LearntVsDesigned}
\end{table*}

\textbf{Comparisons of different contrastive learning frameworks.}
As a plug-and-play and framework-agnostic augmentation algorithm, FocusContrast can work on different contrastive learning frameworks naturally. In this section, we further test over three widely adopted contrastive learning frameworks, namely, MoCo v2~\cite{chen2020improved} (denoted as \textit{MoCo}), \textit{SimCLR}~\cite{chen2020simple}, and \textit{BYOL}~\cite{grill2020bootstrap} in Table \ref{tbl: different frameworks}. The superscript \textit{Learned} indicates that we integrate the proposed FocusContrast with the specific contrastive learning framework.
We also compare to the randomly initialized (\textit{random-init}) and ImageNet pre-trained (\textit{ImageNet}) backbones in the table.

We observe that our FocusContrast improves over conventional random augmentation by +2.72\% (ACC) for \textit{MoCo}, +1.76\% for \textit{SimCLR}, and +3.14\% for \textit{BYOL}, respectively, when 10\% label fraction is available to linear probing. 
A similar and even larger improvement is also achieved when 100\% label fraction is used (i.e., +3.26\%, +3.86\%, +3.50\%, respectively). 
The consistent gains show that our FocusContrast is effective and generalizes well with respect to existing contrastive learning frameworks.
And the performance gains become more significant when compared to non-contrastive pre-training methods (such as \textit{random-init} and \textit{ImageNet}), which are in line with our early experimental results.
\begin{table}[]
\centering
\caption{Linear Evaluation of our method and other contrastive learning frameworks}
\begin{tabular}{lcccc}
\hline
\multicolumn{1}{c}{} & \multicolumn{2}{c}{Label fraction} & \multicolumn{2}{c}{Label fraction} \\
Pretrain method      & \multicolumn{2}{c}{10\%}           & \multicolumn{2}{c}{100\%}          \\
                     & ACC              & MAE             & ACC              & MAE             \\ \hline
random-init          & 38.65            & 1.244           & 39.13            & 1.216           \\
ImageNet             & 43.12            & 0.986           & 44.44            & 0.930           \\ \hline
MoCo                 & 50.84            & 0.746           & 53.68            & 0.674           \\
MoCo$^{Learned}$      & 53.56            & 0.701           & 56.94            & 0.627           \\
SimCLR               & 44.25            & 0.962           & 49.58            & 0.781           \\
SimCLR$^{Learned}$    & 46.01            & 0.878           & 53.44            & 0.697           \\
BYOL                 & 51.57            & 0.755           & 55.31            & 0.685           \\
BYOL$^{Learned}$      & 54.71            & 0.658           & 58.81            & 0.607           \\ \hline
\end{tabular}
\label{tbl: different frameworks}
\end{table}


\textbf{Ablation study on the proposed augmentation operators.}
To further verify the contributions of the three proposed image augmentations, we investigate the performance of linear probing when applying the augmentations individually or in pairs. Three baselines are here for comparison in Table \ref{table:ablation}, including ImageNet pre-trained (\textit{IN pre-trained}), \textit{BYOL}, and BYOL with the optimal augmentation setting exhaustively searched (\textit{BYOL$^{Searched}$}). Our proposed augmentations are added to the baseline of \textit{BYOL}.

Table \ref{table:ablation} shows the linear probing results using 10\%  and 100\% label fractions, respectively. For 10\% label fraction, \textit{focus mask} improves the linear probing accuracy from 51.57\% to 53.62\%. Also, when only \textit{focus crop} is applied, the linear probing accuracy is boosted to 53.44\%. In comparison, \textit{focus cutout} brings in a smaller improvement compared to the other two operators. Similar conclusions are also reported recently in masked image model self-supervised learning~\cite{he2021masked}. We believe that it is because the cutout only removes a small part of the image, which makes the contrastive learning task too easy compared to the other two augmentations. 
As a summary, by adding one or more augmentations to our FocusContrast, the capability of contrastive learning continuously improves as reflected by our experimental results.
\begin{table}[]
\centering
\caption{Ablation study of three proposed augmentation. This table reports the linear evaluation performance. Focus Crop is abbreviated as F Crop and the same applies for Focus Cutout and Focus Mask.}
\begin{tabular}{ccccccc}
\hline
\multicolumn{3}{c}{}                          & \multicolumn{2}{c}{Label fraction}          & \multicolumn{2}{c}{Label fraction}          \\
\multicolumn{3}{c}{Baseline Methods}          & \multicolumn{2}{c}{10\%}                    & \multicolumn{2}{c}{100\%}                   \\
\multicolumn{3}{c}{}                          & ACC                  & MAE                  & ACC                  & MAE                  \\ \hline
\multicolumn{3}{c}{IN pre-trained}            & 43.12                & 0.986                & 44.44                & 0.930                \\
\multicolumn{3}{c}{BYOL}                      & 51.57                & 0.755                & 55.31                & 0.685                \\
\multicolumn{3}{c}{BYOL$^{Searched}$}         & 53.44                & 0.697                & 58.93                & 0.586                \\ \hline
\multicolumn{3}{c}{FocusContrast Opertators} & \multicolumn{1}{l}{} & \multicolumn{1}{l}{} & \multicolumn{1}{l}{} & \multicolumn{1}{l}{} \\
F Crop    & F Cutout  & F Mask    & \multicolumn{1}{l}{} & \multicolumn{1}{l}{} & \multicolumn{1}{l}{} & \multicolumn{1}{l}{} \\ \hline
              &               & $\checkmark$  & 53.62                & 0.699                & 58.81                & 0.607                \\
$\checkmark$  &               &               & 53.44                & 0.696                & 56.94                & 0.620                \\
              & $\checkmark$  &               & 52.16                & 0.723                & 55.89                & 0.652                \\
$\checkmark$  &               & $\checkmark$  & 53.56                & 0.701                & 56.64                & 0.640                \\
$\checkmark$  & $\checkmark$  &               & 53.08                & 0.695                & 56.64                & 0.636                \\
              & $\checkmark$  & $\checkmark$  & 54.12                & 0.674                & 59.43                & 0.603                \\
$\checkmark$  & $\checkmark$  & $\checkmark$  & 54.71                & 0.658                & 59.66                & 0.583                \\ \hline
\end{tabular}
\label{table:ablation}
\end{table}


\section{Conclusion and Discussion}
\label{conclusion}
In this paper, we start by questioning whether the current augmentation strategies popular in general-purpose contrastive learning may hurt the representations learned for medical images. Then we demonstrate that the strong augmentation that works well in natural images cannot fit the pre-training of the medical images, because the small semantic-related areas can be easily corrupted to degrade the learned representations.

Moreover, we propose three learnable augmentation operations guided by radiologists' gaze. Particularly, we use eye-tracker to record the experts' gaze and train a visual attention prediction network to help filter out inappropriate views. The proposed FocusContrast boosts the network's performance significantly in the follow-up tasks compared to default augmentation settings. Our solution requires no extra code or package and works with different contrastive learning frameworks naturally

Our paper shows the feasibility to learn augmentation paradigms from human experts instead of handcrafting them when applying contrastive learning to unseen domains. In the future, we will further extend the gaze-guided approach to more self-supervised frameworks such as masked image models, to guide the network to learn from unlabeled images more efficiently.

There are limitations of this work which we plan to further address in future works. 
First, using a gaze map predicting network to predict the visual attention may potentially slow down the augmentation process. The current workaround is calculating and storing a salience map for each image before contrastive pre-training. 
Then, the \textit{do-while} statement (Algorithm \ref{code}, Line 4) introduces branches thus making the transform hard to be computed in parallel. 
Finally, the knee X-ray image is relatively a simple medical image. 
We are planning to extend this idea to more 2D/3D imaging modalities and clinical applications.

{
\bibliographystyle{IEEEtran}
\bibliography{egbib}
}

\end{document}